# Zero-shot Segmentation of Skin Conditions: Erythema with Edit-Friendly Inversion

Konstantinos Moutselos*[2] [0000-0002-6759-8540] and Ilias Maglogiannis[2] [0000-0003-2860-399X]

[1] University of Piraeus, Dept. of Digital Systems. Piraeus 18534, Greece
`{kmouts;imaglo}@unipi.gr`

**Abstract.** This study proposes a zero-shot image segmentation framework for detecting erythema (redness of the skin) using edit-friendly inversion in diffusion models. The method synthesizes reference images of the same patient that are free from erythema via generative editing and then accurately aligns these references with the original images. Color-space analysis is performed with minimal user intervention to identify erythematous regions. This approach significantly reduces the reliance on labeled dermatological datasets while providing a scalable and flexible diagnostic support tool by avoiding the need for any annotated training masks. In our initial qualitative experiments, the pipeline successfully isolated facial erythema in diverse cases, demonstrating performance improvements over baseline threshold-based techniques. These results highlight the potential of combining generative diffusion models and statistical color segmentation for computer-aided dermatology, enabling efficient erythema detection without prior training data.

**Keywords:** skin erythema, friendly edit inversion, diffusion models.

## 1 Introduction

Erythema (skin redness) is a critical visual indicator in dermatology and is associated with a wide range of conditions, from eczema and rosacea to allergic reactions. Accurate detection and quantification of erythema can assist in diagnosis and treatment monitoring. However, traditional approaches heavily rely on annotated datasets and supervised learning to train segmentation models, which is labor-intensive and may not be well generalized to new conditions or skin tones. Simpler threshold-based techniques (e.g. defining an "erythema index" from color channels) exist but often lack robustness and accuracy, especially for subtle or diffuse redness [1, 2]. There is a need for methods that can detect erythema without extensive training data, i.e. in a zero-shot manner, while handling the variability in lighting and skin appearance.

Generative diffusion models have emerged as powerful tools for image synthesis and editing. Notably, large pretrained diffusion models, such as Stable Diffusion can generate realistic images and perform semantic edits given text prompts [3]. A diffusion model can be inverted to reconstruct a given input image and then used for semantic modifications without retraining using techniques such as edit-friendly inversion [4]. This capability presents an opportunity to create a reference version of a medical image where a certain pathology (e.g., erythema) is removed via editing, while other features remain the same. We hypothesize that comparing an original image with such a synthetic healthy reference will allow unsupervised pixel-wise identification of the pathological features (in this case, red regions).

This research introduces a novel pipeline for zero-shot erythema segmentation, which integrates generative editing and color-based segmentation techniques. Initially, a latent diffusion model generates an image with erythema removed, achieved via prompt-based editing, from an input image featuring erythema. Next, we implement image alignment to ensure the original and the edited images remain spatially aligned. Subsequently, we conduct an analysis in the CIELAB colour space, using the differences in the A* channel to identify areas where there are marked disparities between the original and reference images, indicating the presence of erythema. Our approach does not require any segmentation labels for training; the diffusion model is pre-trained on general images, and the segmentation step is performed unsupervised, thus enabling it to be deployable on new data without the need for retraining.

We validate the method using examples of cases of facial eczema and dermatitis. Our zero-shot approach can highlight erythematous areas that correspond well to clinical observation, even under varied lighting or skin conditions. This study demonstrates the first application of prompt-guided diffusion editing in dermatology, illustrating how generative models can be combined with traditional image analysis to tackle medical segmentation problems in a training-free manner. We discuss the potential of this approach to reduce annotation costs and generalize across different skin conditions exhibiting redness.

## 2   Related Work

**Zero-Shot Segmentation in Vision:** By harnessing prior learned representations, recent research has explored segmentation methods that do not require problem-specific training. Vision-language models such as CLIPSeg (which extends CLIP for segmentation) enable unseen class segmentation via text prompts or embeddings, effectively performing open-vocabulary or zero-shot segmentation tasks [5]. Lüddecke et al. demonstrated that CLIPSeg can produce reasonable masks for objects described by a prompt, without conventional supervised training on those classes. The Segment Anything Model (SAM) [6] is another landmark in this area: Kirillov et al. developed a prompt-based segmentation model trained on a massive dataset to generalize to virtually any object or region given minimal prompts (points, boxes, or text). SAM demonstrates strong zero-shot performance across domains by leveraging powerful image encoder pretraining. Building on SAM, domain-specific adaptations have been proposed; for example, SkinSAM [7] fine-tunes the Segment Anything Model for skin lesion segmentation, yielding improved accuracy on dermatoscopic images by incorporating domain knowledge while retaining SAM's generalization ability. These approaches show the promise of foundation models in segmentation – our work shares the spirit of zero-shot segmentation but focuses on a different strategy, using generative editing instead of direct prompt-based mask prediction.

**Diffusion Models for Medical Imaging and Segmentation:** Denoising diffusion models have become popular for image generation and have also been applied in medical imaging tasks. The Latent Diffusion Model (LDM) by Rombach et al. introduced an efficient way to generate high-resolution images by diffusing in a compressed latent space [3]. Beyond data synthesis, diffusion models have been explored for direct segmentation. Wu et al. proposed MedSegDiff, the first diffusion probabilistic model tailored for medical image segmentation, which conditions the generation process on segmentation maps to achieve high-quality results [8, 9]. Recently, Ni et al. introduced Ref-Diff, a referring image segmentation model that leverages generative diffusion: by using text prompts and the generative prior of Stable Diffusion, Ref-Diff could segment objects described by a phrase in a zero-shot manner, outperforming some supervised models [10]. In the domain of biomedical segmentation without annotations, Hamrani and Godavarty (2025) presented ADZUS (Attention Diffusion Zero-shot Unsupervised System), which uses a self-attention diffusion model to produce segmentations of medical images without training on masks [11]. These works illustrate the versatility of diffusion models: they can act not only as generators but also as unsupervised segmenters by appropriate conditioning or architecture design. Our approach is distinct in that we use diffusion in an image editing role – creating a counterfactual image (with pathology removed) – and then apply a classic segmentation on the difference. This combination of generative and discriminative techniques is, to our knowledge, novel in the context of medical image segmentation.

**Edit-Friendly Inversion in Diffusion Models:** Inversion of diffusion models refers to the process of finding a latent noise vector (or sequence of noise maps) that reproduces a given real image through the generative denoising process. Direct inversion in diffusion is challenging; however, recent advances have made it feasible to achieve nearly exact reconstructions of real images and then perform edits on them [4]. Huberman-Spiegelglas et al. introduced the concept of edit-friendly inversion, proposing to invert images into a non-standard noise space that, while not simply Gaussian noise, allows straightforward manipulations correlating to meaningful image changes. By finding a sequence of diffusion noise maps that perfectly reconstruct the input and have structured correlations across timesteps, allows for simple operations in the noise space, such as shifting objects or changing colors noise space. In text-guided editing, solutions like DiffusionCLIP [12], Prompt-to-Prompt [13], and others, have used approximate DDIM inversions and attention map injections to edit real images. Edit-friendly inversion improves upon these by preserving fidelity without fine-tuning the model, and by maintaining the ability to apply text prompts post-inversion for semantic modifications [4]. In our work, we utilize a similar notion: we invert a dermatological image via a diffusion model (Stable Diffusion) and then guide the generative process with a prompt that excludes erythema, thereby producing an edited image where redness is removed. This approach benefits from the structure preservation that edit-friendly inversion offers – the subject's identity, pose, and lighting remain the same, only the skin redness (a semantic attribute) is altered. To our knowledge, this is the first application of edit-friendly diffusion inversion in a medical context, specifically for skin lesion segmentation.

**Color-Based Dermatological Assessments:** Color analysis has long been used in dermatology for objective assessment of erythema and pigmentation [14]. Traditional techniques often involve handcrafted thresholds or color transformations [15]. For example, earlier studies have used the CIELAB color space – where the *A* channel represents red–green intensity – to quantify erythema by thresholding: pixels with *A* above a certain value are classified as erythematous. Such fixed-threshold methods are straightforward but can be inaccurate, as they do not adapt to varying lighting or baseline skin redness, and they struggle with diffuse or mild erythema. In one rosacea study [1], an automatic threshold approach (as used by a commercial VISIA imaging system) failed to fully segment diffuse facial redness, highlighting the limitation of static thresholds. To improve this, researchers have explored statistical color clustering. Unsupervised algorithms like K-means clustering and Gaussian Mixture Models (GMM) have been applied to skin images to separate pixels into different color- group clusters (for example, separating red lesions from normal skin) based on their color distribution [6]. Overall, these conventional approaches show that color alone contains valuable information for erythema detection, but by themselves they may lack the semantic understanding to differentiate pathological redness from other red elements (like makeup, blood vessels, etc.). Our method bridges this gap by using a generative model to provide semantic guidance: the diffusion-based editing effectively indicates which red areas are abnormal, and then the color clustering can precisely delineate those areas. In essence, we merge the strengths of classical color analysis (precision and simplicity) with those of modern generative AI (semantic context), an approach that to date has not been explored for skin condition analysis.

## 3    Methods

Our proposed pipeline consists of four main stages: (1) Edit-Friendly Inversion to generate erythema-free synthetic reference images using a diffusion model, (2) Image Alignment between original and reference, (3) Skin Region Isolation to focus on skin pixels (optional), and (4) Erythema Segmentation via A* color-space difference. An detailed overview of the whole process is illustrated in Figure 1.

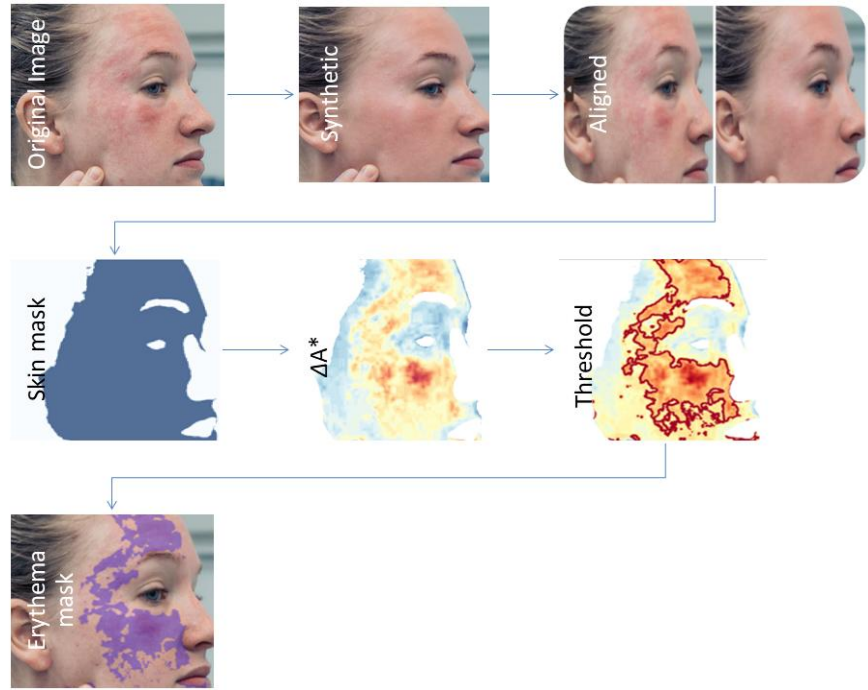

**Fig. 1.** Overview of our zero-shot erythema segmentation pipeline. (a) Original image with erythematous lesions (red patches). (b) Erythema-free reference image generated via edit-friendly diffusion inversion (using a "no redness" prompt). (c) Alignment of the reference to the original (overlaying key facial features). (d) Skin region isolation mask obtained from a pre-trained segmentation model, to restrict analysis to skin areas. (e) ΔA* channel difference between the two aligned images. (f-e). Thresholding and mask overlay. This example uses a woman with facial eczema (image adapted from Cleveland Clinic [16]). The generative model removes the redness on the cheeks and forehead, and after alignment, the difference image in LAB color space can be analyzed for residual red regions.

**Edit-Friendly Inversion & Generative Editing:** We leverage Stable Diffusion (a latent diffusion model) to perform text-guided image editing. Given an input image showing erythema, we first invert it into the model's latent space so that the model can reconstruct the image. Specifically, we use an approach akin to edit-friendly inversion [4]: starting from the input image $I_{orig}$ we find a noise sequence $z_t$ such that the diffusion model, when denoising conditioned on prompt describing the image contents along with the pathology features, recovers $I_{orig}$. Once this inversion is achieved, we modify the conditioning with a prompt that explicitly removes the pathology (erythema). For example, if the input is a face with redness, we use a prompt like "a photograph of a person with clear skin, no redness or rash" as guidance for the diffusion model. The model then generates a new image $I_{ref}$ that is structurally identical to $I_{orig}$ (same person, pose, background) but without the erythema. We take care to preserve other skin features: the prompt only negates redness, and we do not perform excessive diffusion steps to avoid unrelated alterations. In practice, a few iterations (e.g. 50 denoising steps) with classifier-free guidance were sufficient to remove most visible redness while leaving facial details intact. The result of this stage is a reference image $I_{ref}$ that serves as a healthy baseline for the patient. Notably, this editing is done zero-shot – the diffusion model was not trained specifically on these medical edits, but its general knowledge (learned from large image-text data) is being applied to interpret "no redness" and produce a plausible outcome. This step is illustrated in Figure 1a and 1b, where an original image (with erythematous lesions) and the generated erythema-free version are shown side by side.

**Image Alignment:** Although the diffusion model maintains the overall scene, there can be shifts or scale differences between $I_{orig}$ and $I_{ref}$. To enable pixel-wise comparison, we perform precise alignment. We use feature-based alignment with robust estimation: specifically, we detect keypoints in both images (using detectors such as ORB and SIFT for a broad match coverage) and compute descriptors for these points. By matching the descriptors between the original and reference images, we obtain a set of tentative correspondences. We then estimate a transformation (homography) that aligns the reference to the original by filtering out outlier matches using RANSAC. The homography is solved to maximize the number of inliers (consistent keypoint mappings), thus focusing on the unchanged structures, such as facial landmarks. The outcome is a registered pair $I_{orig}$, $I_{ref-aligned}$, where each pixel corresponds to the same location on the subject. After alignment, we crop or mask out any regions that do not perfectly overlap or are non-skin (like edges introduced due to slight scaling differences) to avoid false differences. Figure 1c depicts the result of overlaying $I_{ref}$ onto $I_{orig}$ after alignment. Apart from the absence of redness, the images should now differ minimally.

**Skin Region Isolation**: Before comparing colors, the skin region was isolated to eliminate irrelevant differences or background clutter. For this, we use a pre-trained deep learning model for human skin segmentation. We chose a model available on HuggingFace (based on SegFormer architecture fine-tuned for face parsing) that segments an image into classes including skin, hair, and background. This model (labeled " jonathandinu/face-parsing") was originally trained on CelebAMask-HQ and related datasets and can accurately distinguish skin pixels on faces, even separating them from hair or clothing [17]. This model is applied to the original image $I_{orig}$ to obtain a binary mask $M_{skin}$ of skin vs non-skin. This mask is then applied to both $I_{orig}$ and $I_{ref-aligned}$ setting all non-skin pixels to a neutral value (or simply ignoring them in the analysis). In our experiments (which are mainly frontal face images), the face-hair segmentation was sufficient to capture the areas of interest (facial skin). Figure 1d illustrates the skin mask overlaid on the original image, showing that only the face area is retained for further processing, while the nose and lips are excluded to avoid false positives caused by their natural coloration, which is not reliably indicative of erythema severity.

**Erythema Segmentation via LAB:** With aligned, skin-isolated image pairs, the core of our segmentation is to detect differences in redness. We convert both images from RGB to the CIE La*b color space, which is designed to be perceptually uniform. In LAB, the *A* channel represents the red-green axis (higher $A$ = more red), and *B* represents the blue-yellow axis, while $L$ is the luminance. The difference image $\Delta = I_{orig}^{LAB} - I_{ref}^{LAB}$ (pixel-wise subtraction of LAB values) is computed. We focus on the $\Delta A$ component, where positive values indicate regions in the original image that are redder than those in the synthesized reference. This $\Delta A$ image is evaluated within the previously extracted skin mask.

To isolate erythematous regions, we analyze the histogram of $\Delta A$ values and select a threshold based on its statistical distribution—typically identifying pixels with $\Delta A$ values exceeding the mean by a fixed multiple of the standard deviation (e.g., $\mu + 1.5\sigma$). Pixels with $\Delta A$ values exceeding this threshold are labeled as erythematous. This empirical thresholding method effectively separates inflamed regions from normal skin without the need for probabilistic modeling.

The final erythema mask is directly visualized by overlaying it in translucent color on the original image. This unsupervised and parameter-light method adapts per image, relying solely on visual histogram inspection to determine segmentation thresholds.

## 4    Results and Discussion

We evaluated our zero-shot segmentation approach on three example images that exhibit different erythema manifestations. Figure 2 summarizes the key results on three representative cases: (a) an adult patient with facial eczema (notable red patches on the cheeks and forehead), (b) an infant with atopic dermatitis around the face, and (c) a patient with periocular skin redness. These test images were obtained from public datasets and clinical sources (with sources cited in the figure caption) and they were not part of any model's training data. No ground-truth segmentation masks are available for these images, so we validate the output qualitatively against clinical expectations.

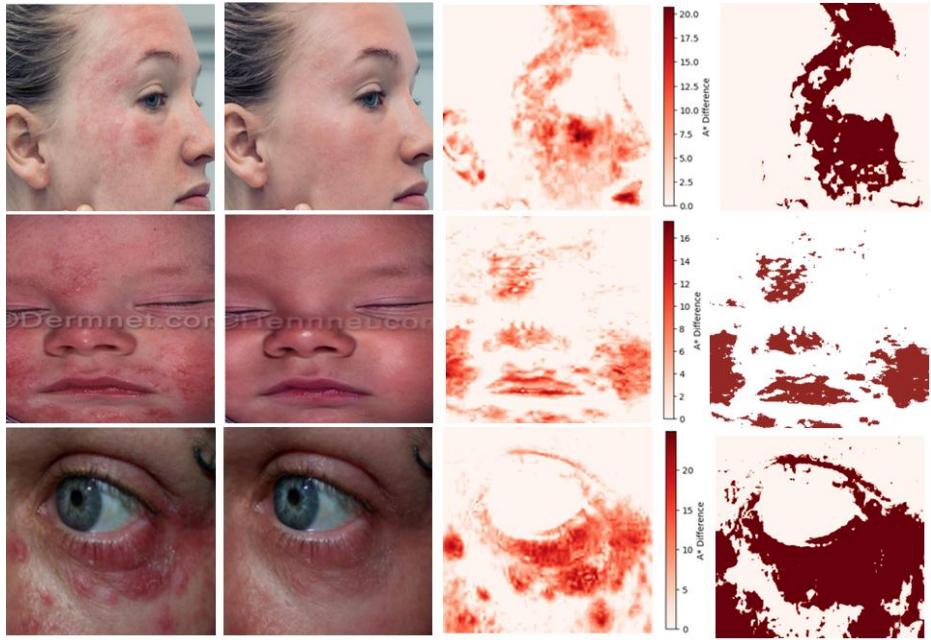

**Fig. 2.** Qualitative results of zero-shot segmentation on various cases of erythema. Original input images (1st col.) and the diffusion-edited reference images with erythema removed (2nd col.). Heatmaps of the difference in red color ($\Delta a$ in Lab space) between original and reference (3rd col.). Final segmentation masks for erythema (4th col.). Sources of original images: (a) Cleveland Clinic [16] (b,c) DermNet New Zealand dataset on Kaggle [18].

**Generative Reference Synthesis:** In all cases, the diffusion model was able to produce a plausible erythema-free image. For the facial eczema image, the edited version had the red rashes on the cheeks and forehead significantly reduced, yielding a more uniform complexion (Figure 2, 2nd column). Importantly, facial features such as the eyes, nose shape, lips, and overall lighting remained consistent between the original and the edited image, confirming that the edit-friendly inversion preserved structural details while removing the targeted attribute (redness). For the baby face with dermatitis, the reference image showed normal skin without the red inflammation, yet the baby's face and expression were unchanged. This demonstrates the model's ability to generalize the "no redness" instruction even to pediatric skin, which has different texture and tone from adult skin. In the ocular case, the generative model cleared the surrounding skin, resulting in a healthier appearance. Overall, these reference images confirm that the diffusion editing step can successfully simulate a treated or healthy skin condition for comparison, even when applied to images captured under markedly different viewpoints, camera distances, and lighting conditions.

**Alignment and Difference Analysis:** Figure 1a and 1b highlight the difference in aspect ratio between the original dermatology image and the rectangular reference image produced through the diffusion-based edit-friendly inversion process. This difference underscores the necessity of a robust alignment step to ensure pixel-level comparability between the two images.

To achieve this, for all three test cases, we applied a homography transformation to spatially align the synthesized reference images to their corresponding originals. When required, central cropping was also performed to remove any padding or border regions introduced during the diffusion process. This procedure, illustrated in columns 1 and 2 of Figure 2, effectively minimized the structural differences between the original and reference images, as quantified by the mean squared error (MSE) metric.

This alignment step is essential because even subtle misalignments could lead to incorrect $\Delta A^*$ computations and, consequently, erroneous erythema masks. By combining homography-based alignment with optional cropping, we ensured that the subsequent difference analysis was both spatially accurate and robust across images with varying aspect ratios and acquisition conditions. Figure 2 (3rd column) visualizes the $a$-channel difference. In the difference image, red areas from the original appear as intense red regions. For example, the cheeks in the eczema case show a strong positive $\Delta a$, whereas the surrounding skin is near zero $\Delta a$.

**Final Segmentation Outputs:** The final segmentation masks outputs are shown in Figure 2 (4th column). In the eczema case (1st row), the mask covers the regions of active dermatitis on the cheeks and on the forehead (the nose and lips regions have not been excluded). In the infant case (2nd row), the method precisely marked the eczematous areas – again aligning well with the known distribution of atopic dermatitis in infants. In the ocular case, the segmentation highlighted the redness around the left eye.

One limitation of our approach is that we rely on the diffusion model's knowledge. If the model does not understand a certain type of lesion or misinterprets the prompt, the reference generation might be suboptimal. In essence, the quality of segmentation is tied to the generative model's capability. As diffusion models improve (especially with more medical image training or fine-tuning), we expect our segmentation results to become even more accurate.

In summary, the results confirm that our method can successfully perform zero-shot erythema segmentation. It effectively exploits the generative model to imagine the same patient without lesions, and uses that as a context to find what's abnormal in the original image. This approach was able to handle different scenarios without retraining or parameter tweaking. We anticipate that as generative models become more commonplace in medical imaging, such approaches could be widely applied to other visible signs (such as bruising, swelling, or even segmenting infections vs normal tissue) by simply tailoring the textual prompt. Our study opens up this avenue, showing feasibility in τhe case of skin erythema. And we suggest a new paradigm for segmentation: "segment by synthesis", where synthesizing a healthy version of an image allows automated segmentation of the pathological features.

## 5   Conclusion

We presented a novel zero-shot approach for segmenting skin erythema by combining diffusion-based edit-friendly inversion with classical color analysis. Without needing any annotated training data, our method effectively identifies red lesions by generating a patient-specific reference image (with no erythema) and pinpointing color differences. In experiments on example cases of facial eczema and atopic dermatitis, the system accurately segmented the erythematous regions, aligning well with clinical expectations. Key contributions of this work include: (1) introducing generative diffusion models as a tool for on-the-fly supervisory signals in medical image segmentation, (2) demonstrating the first use of edit-friendly latent inversion in a dermatology context, and (3) an unsupervised pipeline that combines modern AI (diffusion, transformers for segmentation) with established color clustering methods. The outcome is a flexible diagnostic aid that can be adapted to various skin conditions or imaging scenarios via prompt engineering, avoiding the costly collection of pixel-wise annotations.

We believe this methodology opens up promising directions for computer-aided diagnosis (CAD): for instance, automatically tracking changes in erythema over time by segmenting follow-up photos, or extending to other attributes. Future work will explore applying this zero-shot segmentation framework to clinical image datasets and potentially adding quantitative evaluation by enlisting experts to create reference masks. Additionally, integrating the approach into a single end-to-end network (e.g., a diffusion model that outputs a difference map directly) could be investigated. Nonetheless, the present approach offers a readily implementable solution using existing tools. By significantly reducing dependence on annotated data, it can accelerate deployment of AI in dermatology, especially for rare conditions or those with high variability. In conclusion, our diffusion-based zero-shot segmentation of erythema demonstrates how generative models can be harnessed to solve discriminative tasks in medicine, providing accurate and edit-friendly diagnostic outputs that are aligned with clinical needs.